%% file: main.tex
\documentclass[letterpaper, 10 pt, conference]{ieeeconf}  

\IEEEoverridecommandlockouts                              

\usepackage{cite}
\usepackage{graphics} 
\usepackage{epsfig} 
\usepackage{mathptmx} 
\usepackage{times} 
\usepackage{amsmath} 
\usepackage{amssymb}  
\usepackage{framed}
\usepackage{booktabs}
\usepackage{subcaption}
\usepackage{url}
\usepackage{hyperref}

\usepackage[accsupp]{axessibility}
\usepackage{xcolor}
\usepackage{csquotes}

\newcommand{\my}{TrafficLens\;}

\title{\LARGE \bf
TrafficLens: Multi-Camera Traffic Video Analysis Using LLMs
}

\author{
    Md~Adnan~Arefeen$^{1}$, 
    Biplob~Debnath$^{2}$, 
    and~Srimat~Chakradhar$^{2}$%
    \vspace{1mm}\\
    \small
    $^{1}$NEC Laboratories America, Princeton, NJ, USA\\
    University of Missouri–Kansas City (UMKC), MO, USA\\
    Email: \texttt{aarefeen@nec-labs.com}
    \vspace{1.5mm}\\
    $^{2}$NEC Laboratories America, Princeton, NJ, USA\\
    Emails: \texttt{biplob@nec-labs.com}, \texttt{chak@nec-labs.com}
    \vspace{2mm}\\
    \thanks{
    2024 IEEE 27th International Conference on
Intelligent Transportation Systems (ITSC)
September 24- 27, 2024. Edmonton, Canada © 2024 IEEE. Personal use of this material is permitted.  
    Permission from IEEE must be obtained for all other uses, including  
    reprinting/republishing for advertising or promotional purposes,  
    creating new collective works, for resale or redistribution to  
    servers or lists, or reuse of any copyrighted component of this  
    work in other works.  
    DOI: \href{https://doi.org/10.1109/ITSC58415.2024.10920144}{10.1109/ITSC58415.2024.10920144}}
}

\begin{document}

\maketitle
\thispagestyle{empty}
\pagestyle{empty}

\input{sections/abstract}
\input{sections/intro}

\input{sections/relatedwork}
\input{sections/motivation}
\input{sections/method}

\input{sections/evaluation}

\input{sections/ablation}

\input{sections/conclusion}

\bibliography{ref}
\bibliographystyle{IEEEtran}

\end{document}

%% file: sections/abstract.tex
\begin{abstract}
Traffic cameras are essential in urban areas, playing a crucial role in intelligent transportation systems. Multiple cameras at intersections enhance law enforcement capabilities, traffic management, and pedestrian safety. However, efficiently managing and analyzing multi-camera feeds poses challenges due to the vast amount of data. Analyzing such huge video data requires advanced analytical tools. While Large Language Models (LLMs) like ChatGPT, equipped with retrieval-augmented generation (RAG) systems, excel in text-based tasks, integrating them into traffic video analysis demands converting video data into text using a Vision-Language Model (VLM), which is time-consuming and delays the timely utilization of traffic videos for generating insights and investigating incidents.
To address these challenges, we propose \my, a tailored algorithm for multi-camera traffic intersections. \my employs a sequential approach, utilizing overlapping coverage areas of cameras. It iteratively applies VLMs with varying token limits, using previous outputs as prompts for subsequent cameras, enabling rapid generation of detailed textual descriptions while reducing processing time. Additionally, \my intelligently bypasses redundant VLM invocations through an object-level similarity detector. Experimental results with real-world datasets demonstrate that \my reduces video-to-text conversion time by up to $4\times$ while maintaining information accuracy.

\end{abstract}

%% file: sections/intro.tex
\section{Introduction}

Traffic cameras have become ubiquitous in urban environments, with many cities installing hundreds to thousands of them. These cameras serve the purpose of continuously capturing video footage of traffic scenarios. The collected videos are then systematically stored for post-analysis. This extensive archive of video data offers city planners and transportation authorities a valuable resource for extracting insights, conducting investigations, preventing potential disasters, and addressing various inquiries related to traffic management. The sheer volume of archived traffic data is immense. For instance, a city with approximately one thousand of these cameras may accumulate as much as 230 terabytes of video data each month~\cite{etl_vid_stream}. In a multi-camera setup that captures the same scene from different angles, the volume of traffic data can double or even quadruple than single camera video feeds. Analyzing such a vast amount of video feeds from traffic cameras is essential for tasks such as traffic monitoring, congestion management, and incident detection. However, this process demands a comprehensive understanding of the information embedded within the videos, highlighting the necessity for advanced analytical tools and methodologies~\cite{mittal2023ensemblenet,llmvideosurvey,shibata2023listening}.

In the domain of traffic video analysis, processing user queries through natural language processing enables direct interaction with video content. Large Language Models (LLMs), such as ChatGPT~\cite{bubeck2023sparks}, have excelled in text-based interactions but face limitations when addressing data not encountered during training. The Retrieval-Augmented Generation (RAG)~\cite{rag} approach has been widely adopted to augment LLMs with the ability to integrate unseen data. However, traditional RAG systems are designed to handle textual data, posing challenges when dealing with non-textual formats like videos or images common in traffic monitoring. Addressing this gap requires enhancing RAG systems with capabilities to convert these media into text-compatible formats. 


\begin{figure*}[!htpb]
    \centering
    \includegraphics[width=0.8\linewidth]{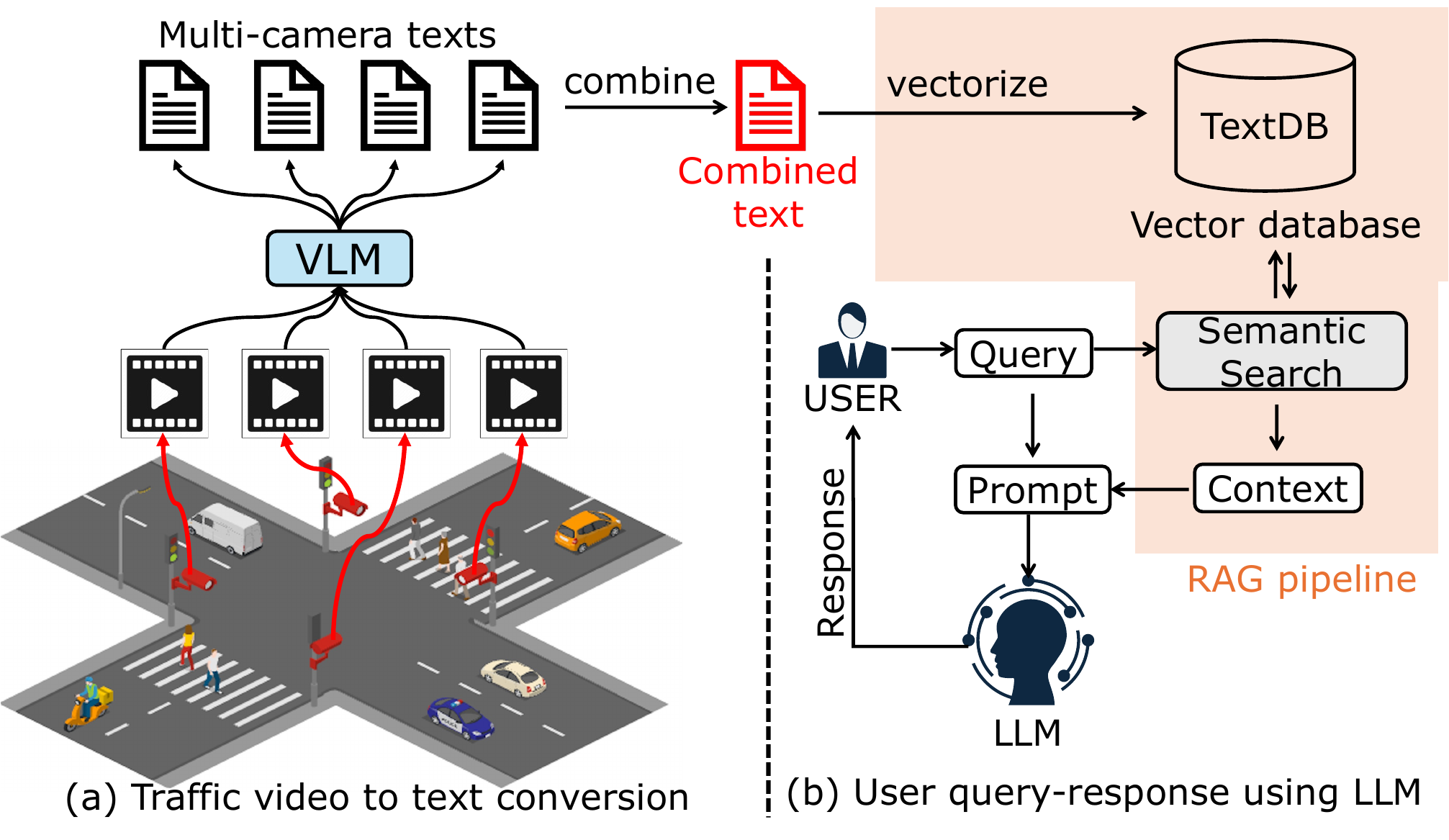}
    \caption{An overview of the RAG-based  traffic video analysis system. It operates in two phases: a) the multi-camera videos is initially converted into a text document, the combined texts from all documents are then chunked into smaller parts and stored in a vector database; b) queries are answered using a Large Language Model (LLM), leveraging query and contextual information retrieved from the vector database through semantic search.}
    \label{fig:video-rag-new}
\end{figure*}

Typically, Vision-Language Models (VLMs)~\cite{dinh2024trafficvlm,tian2024drivevlm,dong2024internlm} are employed to transcribe traffic videos into text, which is then processed through a RAG-based system using Large Language Models. While crucial, this conversion process often becomes a bottleneck, particularly when dealing with a large corpus of videos. It leads to delays in utilizing traffic videos promptly for generating actionable responses. This challenge is further compounded by the deployment of multiple cameras at strategic points within a traffic intersection, as depicted in Figure~\ref{fig:video-rag-new}.

The multi-camera setup at a traffic intersection is designed with redundancy in mind. Each camera complements the others by capturing events from different angles and perspectives. In situations where one camera may fail to capture a particular event due to obstructions or limitations in its field of view, neighboring cameras are strategically positioned to fill in these gaps. This coordinated arrangement minimizes the risk of incidents going unnoticed, as there is a high probability that any event overlooked by one camera will be captured by another.

Figure~\ref{fig:video-rag-new} illustrates a RAG-based system designed to address questions related to multi-camera traffic videos. To initiate the process, the video RAG system converts the videos into text, dividing it into non-overlapping clips. Each clip undergoes analysis by a Vision-Language Model (VLM)~\cite{vlog, mmvid, arefeen2024irag,arefeen2024vita}, recording the output in text format. The video feed from each camera is processed by a VLM. The outputs from all VLM cameras are combined to generate the final textual description of events captured at the traffic intersection clip by clip basis. Collating text information from all clips generates a long document for the multi-camera videos, which is then segmented into chunks. Each chunk is embedded into a vector by an embedding model and subsequently stored in a vector database. Once the video-to-text conversion is complete, upon receiving a query, the video RAG system embeds the query, and conducts a semantic search using the embedding vectors of chunks to retrieve relevant chunks from the vector database. These retrieved chunks form the context, which is combined with the query to generate a prompt. Finally, the prompt is fed into a Large Language Model (LLM) to generate a response for the query.

While a RAG-based system for traffic videos can address a wide range of queries, a major challenge lies in the time required for a VLM model to generate textual descriptions from video clips. For instance, processing a frame using the InternLM-XComposer2~\cite{dong2024internlm}, with a maximum token limit of 64, takes an average of 3.8 seconds on a server equipped with an NVIDIA GeForce RTX 3090 GPU. Thus, it would take more than a day for this VLM model to analyze a 24-hour traffic video from one camera and generate text descriptions, assuming it processes one frame every three seconds. Now, if converting 24-hour traffic videos from one camera takes one day, and an intersection is monitored using four cameras, then it would take more than 4 days just to extract text from the video footages before starting analysis through LLMs. This poses a significant challenge for various applications, such as preventing law enforcement agencies from conducting timely analyses of criminal incidents captured in the traffic videos.


Vision-Language Models (VLMs) essentially function as predictive models for the next token in a sequence. They take input in the form of images or video clips, along with a text prompt, and produce text as output. Within the model, inputs are tokenized initially, and these tokens can be processed in parallel, meaning that the length of the input tokens does not significantly affect latency. However, the length of the generated output significantly affects latency due to sequential token generation.  The size of the generated output is influenced by the prompt. Moreover, VLMs offer a maximum output token limit parameter to regulate the amount of generated information, thereby controlling inference speed. In this paper, our aim is to accelerate the video-to-text conversion process by refining the text prompt and adjusting the maximum token limit parameter of VLMs, leveraging the unique attributes of multi-camera setups deployed at traffic intersections.

To accomplish this goal, we propose \my, an innovative algorithm designed to quickly produce textual descriptions of video clips using VLM models for monitoring traffic intersections equipped with multiple cameras. It employs a sequential approach, capitalizing on the overlapping coverage areas of the cameras at these intersections. Initially, \my applies VLM to the video clip from one camera, prompting it to generate detailed descriptions while employing a higher token limit. Subsequently, \my utilizes the resulting output as a prompt for the next camera, instructing VLM to include additional details not initially covered, while enforcing a lower token limit. This iterative process continues for subsequent cameras, with each iteration incorporating further details from previous cameras and reducing the token limit. Furthermore, it can bypass subsequent VLM calls when it detects a high degree of similarity among the video feeds from different cameras. 

In summary, we make the following contributions:

\begin{itemize}
    \item We present \my, a novel algorithm for accelerating video-to-text conversion using VLMs for traffic intersections equipped with multiple cameras. It employs a higher token limit for the first camera to extract comprehensive text, while applying a lower token limit for subsequent cameras to capture objects undetected by the preceding cameras.

    \item \my utilizes intelligent prompt engineering to adjust the token limit in subsequent camera videos during the video-to-text conversion process. Additionally, it reduces conversion time by eliminating redundant clips from subsequent cameras through object-level similarity detection.

    \item Our experimental evaluation using the StreetAware dataset~\cite{streetaware}, which covers traffic intersections in New York City, demonstrates that \my can accelerate the video-to-text conversion time by up to $4\times$ while maintaining information accuracy.

\end{itemize}

%% file: sections/relatedwork.tex
\section{Related Work}


Vision-Language Models (VLMs) represent a crucial advancement in the fields of Natural Language Processing (NLP) and Computer Vision (CV). By integrating textual and visual data, they enable a deeper understanding of multimodal content. Through VLMs, transportation systems are able to deeply understand real-world environments, thereby improving driving safety and efficiency. To build an interactive system for advanced traffic video understanding, along with VLMs, large language models are also necessary to query on textual description of traffic video data. Using retrieval augmented generation (RAG)~\cite{rag,arefeen2024irag,arefeen2023leancontext}, video data can be incorporated as text to execute interactive query by the users. A survey is conducted by Zhou et al.~\cite{vlm_survey_its} to explore the application of VLMs in intelligent transportation systems. VLMs concerning traffic data are primarily categorized into three types: Multimodal-to-Text (M2T)~\cite{fu2024drive}, Multimodal-to-Vision (M2V)~\cite{wang2023drivedreamer}, and Vision-to-Text (V2T)~\cite{liu2023traffic,dinh2024trafficvlm}. While V2T models take images or videos as inputs and generate textual descriptions as outputs, the M2T models take inputs in the form of image-text or video-text pairs. They analyze both the visual and textual components and generate textual descriptions as output.
For example, when provided with an image depicting traffic congestion alongside its corresponding textual description, an M2T model can produce a detailed textual narrative of the scene, encompassing elements such as traffic flow, weather conditions, and levels of road congestion.

In this paper, our focus is on designing a large-scale traffic video analysis system utilizing multimodal-to-Text (M2T) models and Large Language Models (LLMs). Current M2T models operate on image or video clips spanning several seconds. However, we are dealing with longer videos. Therefore, we use M2T models as foundational components to process longer videos by dividing them into smaller clips. Nonetheless, these models take longer to process images or video clips into textual form. Hence, we leverage the distinctive features of multi-camera setups deployed at traffic intersections to expedite the text conversion process. While some works exist related to the multi-camera setup~\cite{Shim_2021_CVPR,Multi-camera-itsc-2017}, they primarily focus on object detection and tracking. In contrast, our focus lies in generating a textual description by combining the information observed by individual cameras.

%% file: sections/motivation.tex
\section{\my}
\subsection{Motivation}\label{sec:motivation}

\begin{figure}[!htbp]
\centering
\begin{subfigure}[b]{0.42\linewidth}
    \includegraphics[width=\linewidth]{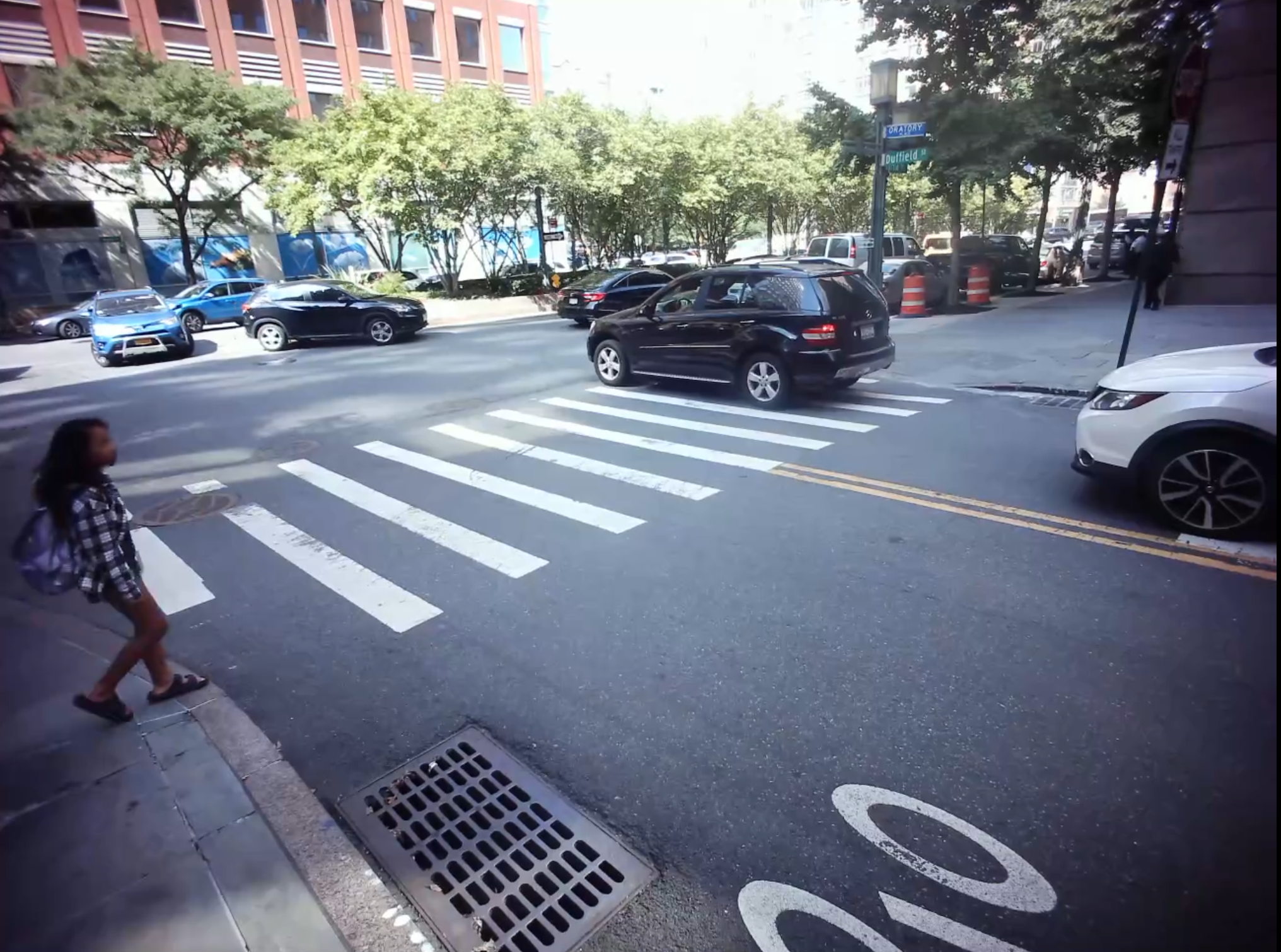}
    \caption{Left camera}
    \label{fig:sub1}
\end{subfigure}
\hfill
\begin{subfigure}[b]{0.48\linewidth}
    \includegraphics[width=\linewidth]{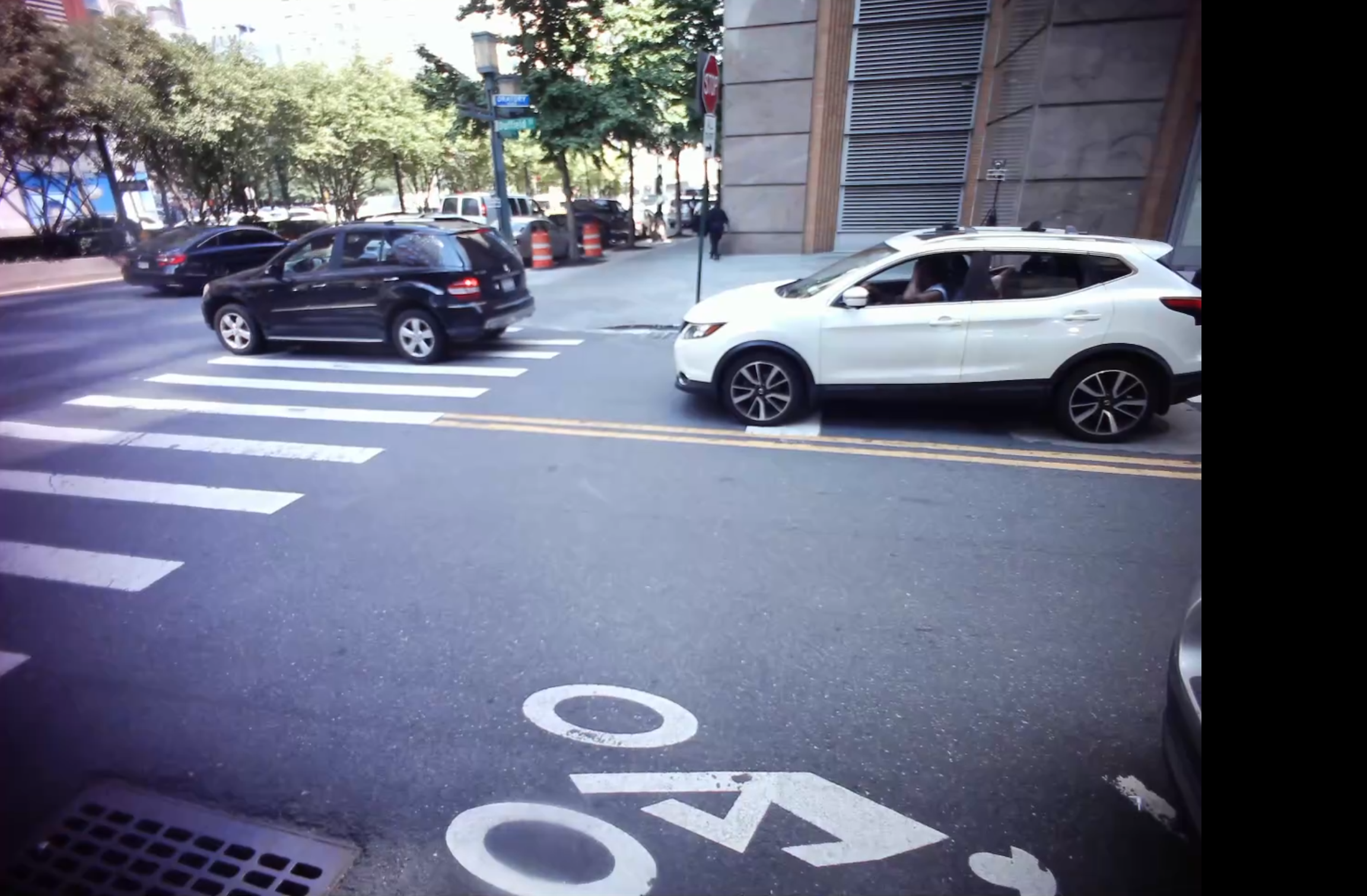}
    \caption{Right camera}
    \label{fig:sub2}
\end{subfigure}
\caption{Left and right camera view in StreetAware~\cite{streetaware} dataset. Unique information of a person wearing a plaid shirt is absent from right camera view.}
\label{fig:motivation-multi-camera}
\end{figure}

Multiple cameras provide broader coverage of the traffic intersection, reducing blind spots. If one camera fails, others can still capture necessary footage. Figure~\ref{fig:motivation-multi-camera} shows views from two cameras at a traffic intersection~\cite{streetaware}, capturing simultaneous moments. This multi-camera setup reveals additional details, such as a person not visible from the right camera, while both cameras capture the white and black cars. To enhance efficiency in multi-camera traffic video analysis, merging overlapping information with unique details from each camera only once can significantly reduce the video-to-text conversion process. 

We have identified three key strategies to accelerate video-to-text conversion in a multi-camera traffic video analysis system utilizing large language models:

\begin{enumerate}

    \item \textbf{Efficient Use of VLMs:} Unlike traditional machine learning models with fixed processing times, the inference time of a Vision-Language Model (VLM)~\cite{dinh2024trafficvlm,tian2024drivevlm,dong2024internlm} varies depending on the complexity of the prompt and the size of the generated output. Adjusting the \textit{maximum token limit} and refining prompts can help reduce processing times.

    \item \textbf{Addition of Unique Information Across Cameras:} Using specific prompts such as  ``Describe the undetected objects only" instead of a general narrative such as ``Compose a descriptive narrative" can prevent redundant processing across cameras, further reducing conversion time as VLM inference depends on the number of output tokens.

    \item \textbf{Elimination of Overlapping Information:} A similarity detector can prevent reprocessing video footage from subsequent cameras that duplicate information from earlier feeds. Additionally, it can help reduce the hallucination problem in VLMs~\cite{bai2024hallucinationmultimodallargelanguage}.

\end{enumerate}

\begin{figure}[!htpb]
    \centering
\includegraphics[width=0.45\linewidth]{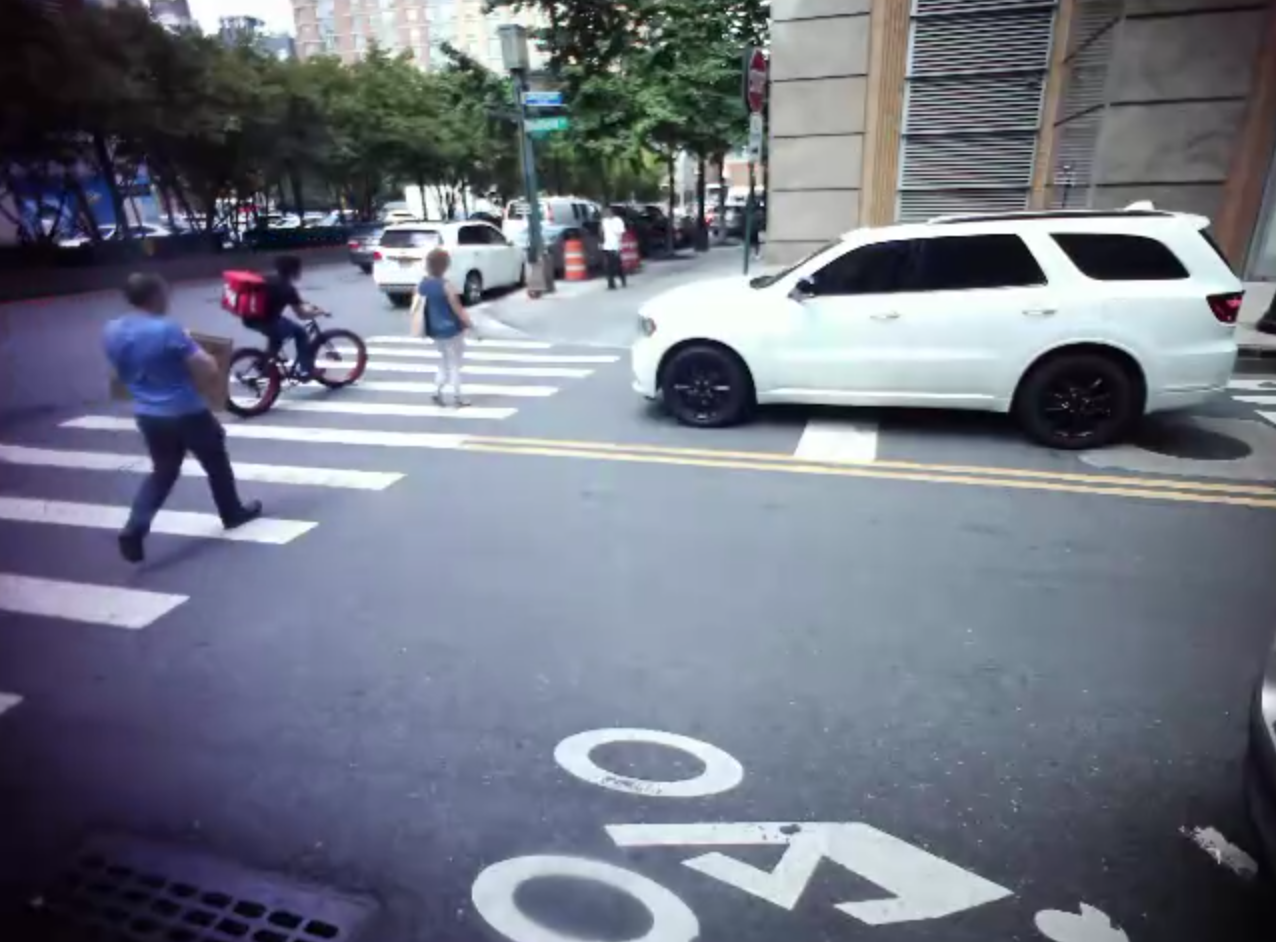}
    \caption{A sample image from the StreetAware~\cite{piadyk2023streetaware} dataset.}
    \label{fig:sample-img}
\end{figure}

To illustrate the effect of the \textit{maximum token limit} parameter (which controls the amount of text generated by the VLM models) on inference time, let us examine the image depicted in Figure~\ref{fig:sample-img}, which shows a street scene in New York City sourced from the StreetAware dataset~\cite{piadyk2023streetaware}.

We generate textual descriptions using the InternLM-Xcomposer2 VLM model (with 1.8 billion parameters)~\cite{dong2024internlm} with the prompt ``Compose a descriptive narrative" by setting the maximum output token limit parameter to 128. It generates the following output:

\begin{framed}\label{text}
\noindent\textbf{InternLM-Xcomposer2-1.8b:}\\
Output Tokens: \textcolor{green}{16}, \textcolor{cyan}{32}, \textcolor{red}{64}, \textcolor{blue}{128}\\
\noindent \textcolor{green}{The image captures a bustling city intersection, where a man in a blue shirt} \textcolor{cyan}{is crossing the street, a bicycle rider in the background, and a white SUV }\textcolor{red}{parked on the side. The intersection is marked by a crosswalk, and a bike lane is also visible. The backdrop of the scene is a mix of urban }\textcolor{blue}{architecture, including buildings and trees, and a clear blue sky.}
\noindent
\end{framed}

The total number of output tokens generated above is 77. Different colors indicate the additional details added as the token limit increases. It is important to note that the maximum output token limit serves as guidance for the VLM to control the length of the generated text. However, depending on the input image and prompt, it may produce shorter outputs.  

To demonstrate the relationship between inference latency and output token size in VLMs, we conducted experiments with various maximum output token limits: specifically, 16, 32, 64, 128, and 256.  We tested these limits on VLM models including InternLM-Xcomposer2 (with 1.8 billion and 7 billion parameters~\cite{dong2024internlm}, LLAVA-1.5-7b model~\cite{liu2023llava}, and MobileVLM with 1.7 billion parameters~\cite{chu2024mobilevlm}.
This range of limits enables us to observe how each model responds to different token limit parameters.

\begin{figure}[!htpb]
    \centering
    \includegraphics[width=\linewidth]{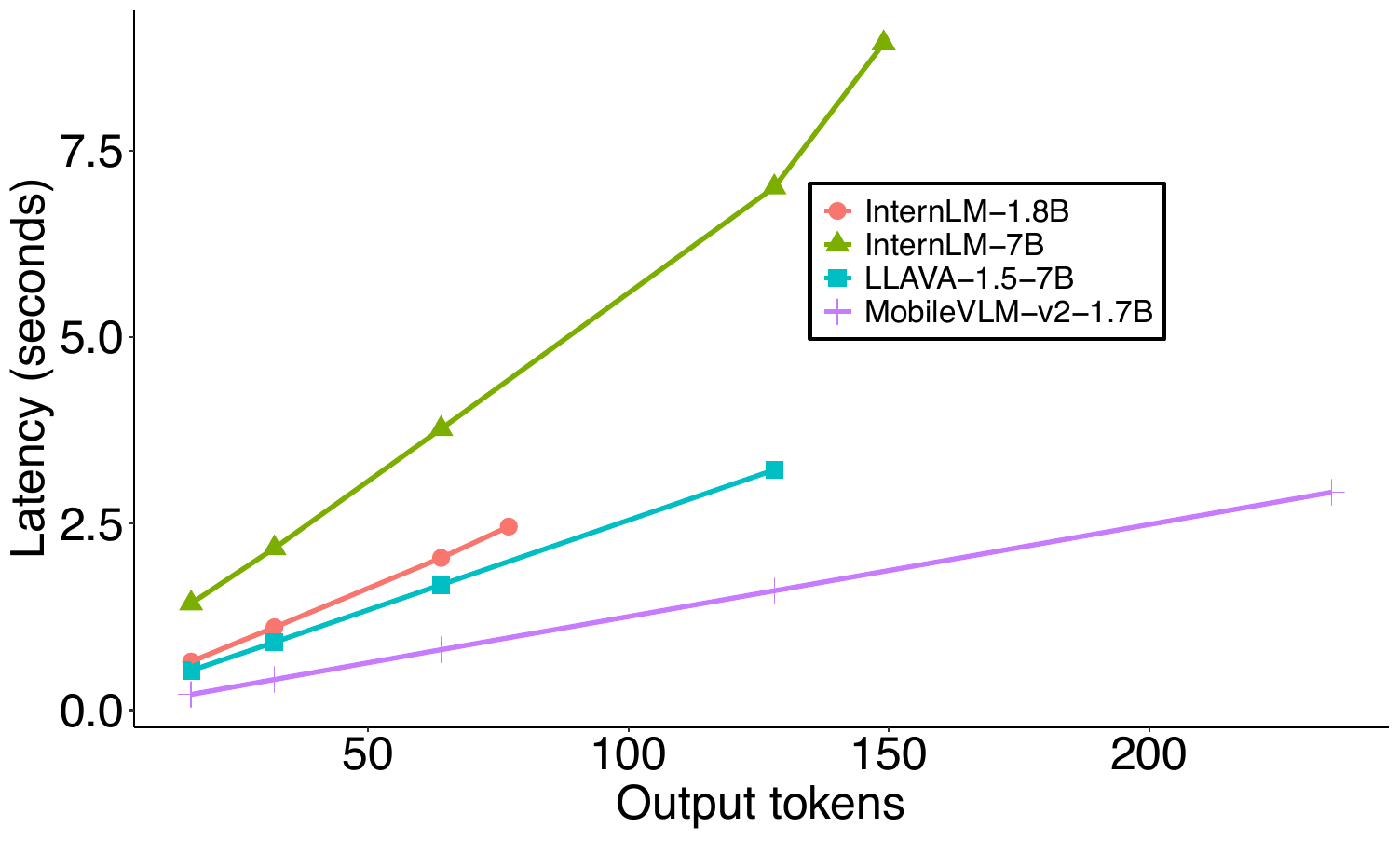}
    \caption{Latency of VLM output generation increases with the increase in number of output tokens.}
    \label{fig:output_length_vs_latency}
\end{figure}

In Figure~\ref{fig:output_length_vs_latency}, we observe a clear relationship between the maximum token limit and the inference latency across various VLMs. As the maximum token limit increases, the time taken for inference also consistently increases across all VLMs. This trend indicates that as VLM models are tasked with generating longer textual descriptions, they require more time to complete the inference process. These observations motivate us to build an interactive system, TrafficLens, that can efficiently convert multi-camera videos to text for further analysis with large language models.

%% file: sections/method.tex
\begin{figure*}[!htbp]
\centering
\begin{subfigure}[b]{0.49\linewidth}
    \includegraphics[width=\linewidth]{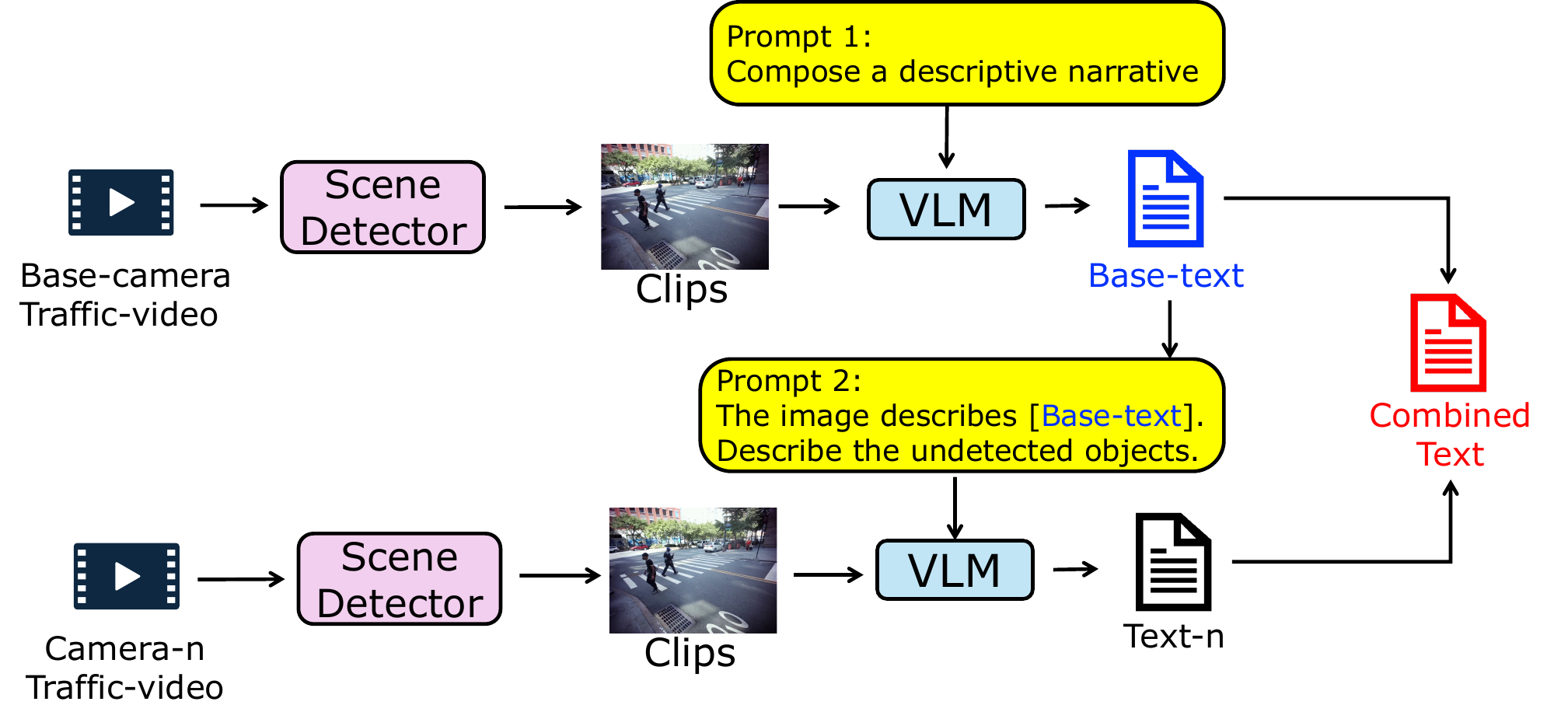}
    \caption{Different prompts are employed to incrementally capture textual information across cameras.}
    \label{fig:int_ingest}
\end{subfigure}
\hfill
\begin{subfigure}[b]{0.49\linewidth}
    \includegraphics[width=\linewidth]{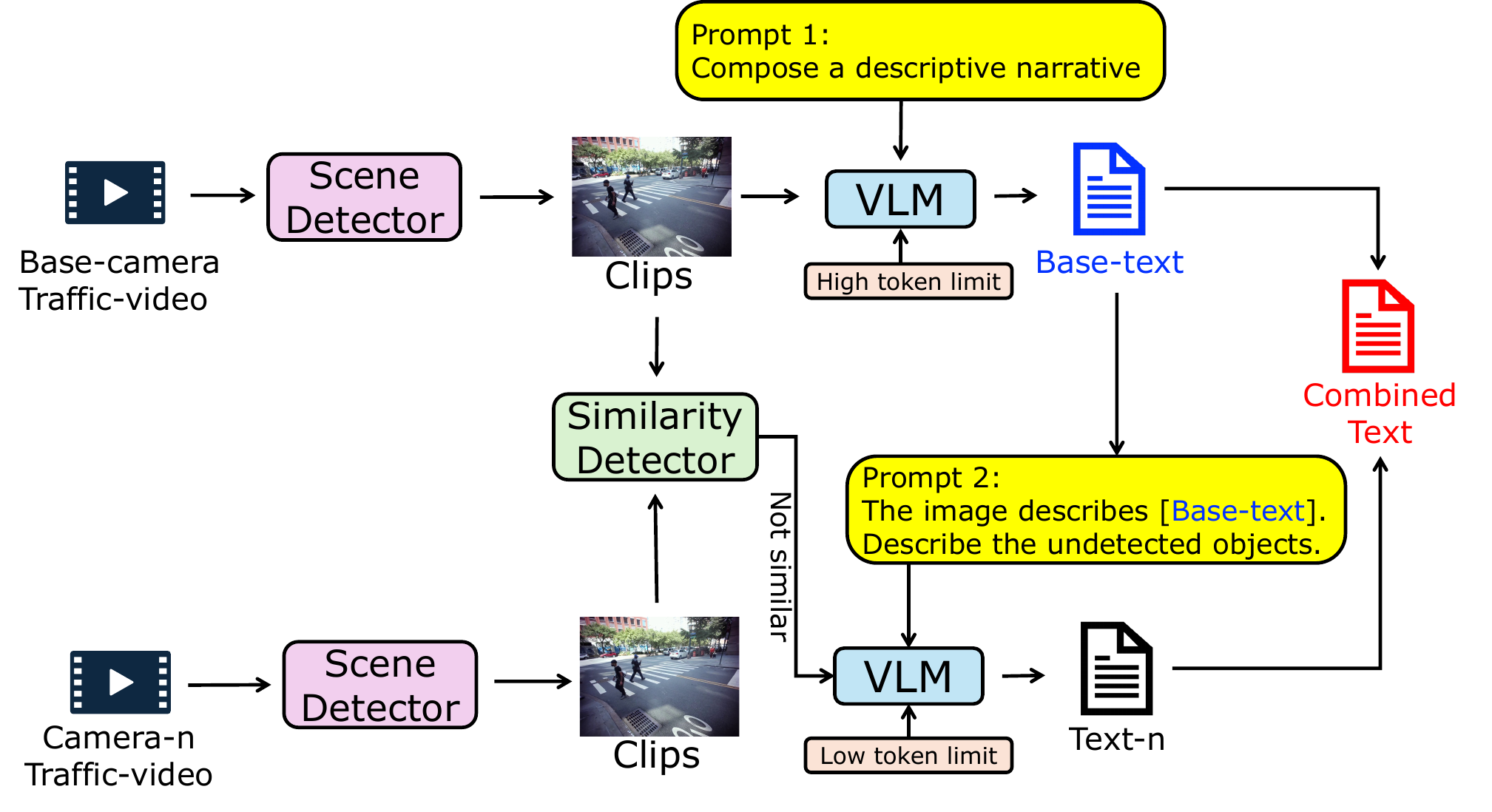}
    \caption{Token limits are adjusted, and a clip similarity detector is utilized to further reduce the time required for textural information extraction.}
    \label{fig:final_ingest}
\end{subfigure}
\caption{Accelerated video-to-text conversion workflow of TrafficLens.}
\label{fig:test}
\end{figure*}

\subsection{Proposed Method}

Multiple cameras covering the same view from different positions provide unique perspectives. While there is some overlap in the scenes captured, each camera may also record objects that are only visible from its specific vantage point. Consequently, in a multi-camera setup at a traffic intersection, extracting information independently from each camera can lead to the accumulation of redundant data, which in turn impacts the efficiency of video ingestion. To eliminate redundancy from accumulating information from all cameras, \my implements an incremental approach. It first considers a camera as base and extracts the scene information.

\subsubsection{Base-camera Ingestion}
\my starts with selection of a base camera video ingestion. At first, \my runs a scene detector to identify the non-overlapping clips from the video corpus. With an open source vision language model (VLM), \my runs the following prompt to extract the details
:

\begin{framed}
\noindent\textbf{Prompt 1:}\\
Compose a descriptive narrative.
\end{framed}

\subsubsection{Camera-n ingestion}
After completing the ingestion from the base camera, \my obtains the base text for each clip from the base camera. Subsequently, \my  employs Prompt 2 to delve deeper and extract additional details from the clips of the next camera with the same time frame as base-camera. In this way, unique information from all cameras will be covered. Since multiple cameras cover the same scene from various angles, this approach leverages the unique positioning of each camera to enrich the overall context and understanding of the scenario.

\begin{framed}
\noindent\textbf{Prompt 2:}\\
The image describes [\textcolor{blue}{Base-text}]. Describe the undetected objects.
\noindent
\end{framed}


Figure~\ref{fig:int_ingest} depicts the
accelerated video-to-text conversion workflow through prompt refinement of TrafficLens. Varied prompts are employed to progressively capture textual information across cameras. Prompt 2 is specifically designed to compel the VLMs to extract information solely about objects that were not detected by previous camera feeds. This approach significantly reduces redundancy in subsequent camera feeds. 

We have observed that when using Prompt 1, the base camera text generally captures most of the necessary information. Consequently, when Prompt 2 is applied to subsequent cameras, it often redundantly identifies objects that have already been detected. An example is shown below, where repeated information is indicated in red.

\begin{framed}
   \noindent \textcolor{blue}{\textbf{Base-text}}: The image captures a moment on a city street, where \textcolor{red}{a pedestrian is seen crossing the road}. The pedestrian, dressed in a black shirt and shorts, carries a backpack, suggesting they might be a student or a commuter. The street is marked with a bike lane, indicating a focus on eco-friendly transportation. The background reveals \textcolor{red}{a building with a large window, and a car parked nearby}, adding to the urban setting. The perspective of the image is from the side of the road, providing a clear view of the pedestrian and the surrounding environment. \\
   
   \noindent \textcolor{black}{\textbf{Camera-n text}}: The undetected objects in the image include \textcolor{red}{a building with a large window, a car parked nearby, and a pedestrian crossing the road.}
\end{framed}


Based on this observation, \my introduces two additional techniques to enhance the ingestion capability of VLMs for the multi-camera setup at the traffic intersections:  Clip Similarity Detector and Reduced Token Limit. Figure~\ref{fig:final_ingest} illustrates the workflow for the enhanced video-to-text ingestion process of TrafficLens. \\


\noindent\textbf{Clip Similarity Detector.} In most multi-camera setups, there is an overlapping region where the same objects are captured by all cameras. Additionally, during periods of no traffic (i.e., when roads are empty), all cameras capture no objects, thus sharing similar information. Taking this into account, TrafficLens implements a similarity detector to identify similarities in multi-camera scenes. It uses object-level similarity as a metric for similarity detection, employing the Intersection over Union (IoU) score for quantification. When the IoU score exceeds a specified threshold, the VLM is not invoked for that clip to avoid generating redundant textual information already obtained from the base camera. This approach enables a more refined analysis of visual data across different camera feeds.

Eliminating similar clips also contributes to reducing the hallucination problem~\cite{bai2024hallucinationmultimodallargelanguage} in VLMs. Specifically, when Prompt 2 is invoked to generate additional descriptions about undetected objects, the VLM may hallucinate (i.e., produce fabricated information). By avoiding the use of the VLM with similar clips, \my not only reduces video-to-text conversion time but also mitigates the hallucination problem. \\


\noindent\textbf{Reduced Token Limit.} We discuss in Section~\ref{sec:motivation} that the latency involved in converting video feeds to text is influenced by the number of output tokens produced by VLMs. To reduce ingestion time, we can limit the number of tokens generated during this conversion process. For a multi-camera setup in \my, when calling the VLM $n$ times for $n$ cameras, we first generate a detailed description of the base camera feed using Prompt 1 with a higher token limit. Subsequently, we use a tailored Prompt 2 to generate additional information for other camera feeds, specifically targeting details that may have been missed by the base feed. When invoking Prompt 2, \my reduces the token limit. This strategy significantly reduces the time required to process multi-camera video feeds at traffic intersections by cutting down on the number of generated tokens and eliminating redundant information.

%% file: sections/evaluation.tex
\section{Evaluation}
\label{sec:eval}

For evaluation of \my, we consider StreetAware~\cite{piadyk2023streetaware} dataset that focuses on observing pedestrian movement in intersections using REIP sensors. The description of the dataset is shown in Table~\ref{tab:dataset}. We consider videos of two cameras from the StreetAware dataset from two camera positions i.e. left, and right. Each video is 46 minutes 44 seconds long.

\begin{table}[!htpb]
    \centering
    \caption{Description of datasets with ingestion time}
    \begin{tabular}{cccc}
    \toprule
      Camera & Model & Video Duration  & Ingestion time \\
       &  & (hh:mm:ss) & (hh:mm:ss) \\
    \toprule
        Left & InternLM-1.8B & 00:46:44 & 00:28:54 \\
         & LLAVA-7B & 00:46:44 & 00:30:22 \\
    \midrule
      Right & InternLM-1.8B & 00:46:44 & 00:27:19\\
       & LLAVA-7B & 00:46:44 & 00:30:57\\
    \bottomrule
    \end{tabular}
    
    \label{tab:dataset}
\end{table}

\begin{table}[!htpb]
    \centering
    \caption{Token generation statistics under maximum token limit}
    \resizebox{\linewidth}{!}{%
    \begin{tabular}{cccccc}
    \toprule
      \begin{tabular}[c]{@{}c@{}}Camera  \end{tabular}  & Model & \begin{tabular}[c]{@{}c@{}}Max.  \\token \\limit\end{tabular} &\begin{tabular}[c]{@{}c@{}}Max.  \\output \\tokens\end{tabular}  & \begin{tabular}[c]{@{}c@{}}Min.  \\output \\tokens\end{tabular}& \begin{tabular}[c]{@{}c@{}}Avg.  \\output \\tokens\end{tabular}\\
    \toprule
        Left & InternLM-1.8B & 256 & 254 & 53 & 113.14 \\
         & LLAVA-7B & 256 & 213 & 77 & 142.18 \\
    \midrule
      Right & InternLM-1.8B & 256 & 231 & 53 & 105.65\\
       & LLAVA-7B & 256 & 214 & 77 & 144.82\\
    \bottomrule
    \end{tabular}}
    
    \label{tab:token-detials}
\end{table}
\begin{table*}[!htpb]
\centering
\caption{Comparison of ingestion time between \my and the baseline approach.}
\resizebox{\textwidth}{!}{%
\begin{tabular}{ccc|ccccc}
\toprule
No of cameras & Total video duration & Model         &          &                                                         &         Total ingestion time (hh:mm:ss)   &  &\\ \cmidrule{1-8} 
 &  &          &      \multicolumn{1}{l|}{}    &                                                         &         & \multicolumn{1}{l}{\my}  &\\ \cmidrule{5-8} 

             &                      &               & \multicolumn{1}{c|}{Baseline} & \begin{tabular}[c]{@{}c@{}}$T_r=80$, $T_\ell =32$\\ w/o Similarity Detector\end{tabular} & \begin{tabular}[c]{@{}c@{}}$T_r=80$, $T_\ell =32$\\ w Similarity Detector\end{tabular} & \begin{tabular}[c]{@{}c@{}}$T_r=50$, $T_\ell =32$\\ w/o Similarity Detector\end{tabular} & \begin{tabular}[c]{@{}c@{}}$T_r=50$, $T_\ell =32$\\ w Similarity Detector\end{tabular}\\ \cmidrule{4-8}
2            & 01:33:28             & InternLM-1.8B & \multicolumn{1}{c|}{00:56:13} & 00:28:32                                                                      & 00:25:16  & 00:21:09 & 00:18:07   \\ \cmidrule{3-8}
             &                      &    LLAVA-1.5v-7B        &        \multicolumn{1}{c|}{01:01:19}    &                                                                            00:25:42    &            00:22:21  &00:19:44 & 00:16:17\\ \bottomrule
\end{tabular}%
}
\label{tab:ingestion-time-comparison}
\end{table*}
\begin{figure*}[!htpb]
    \centering
    \includegraphics[width=\textwidth]{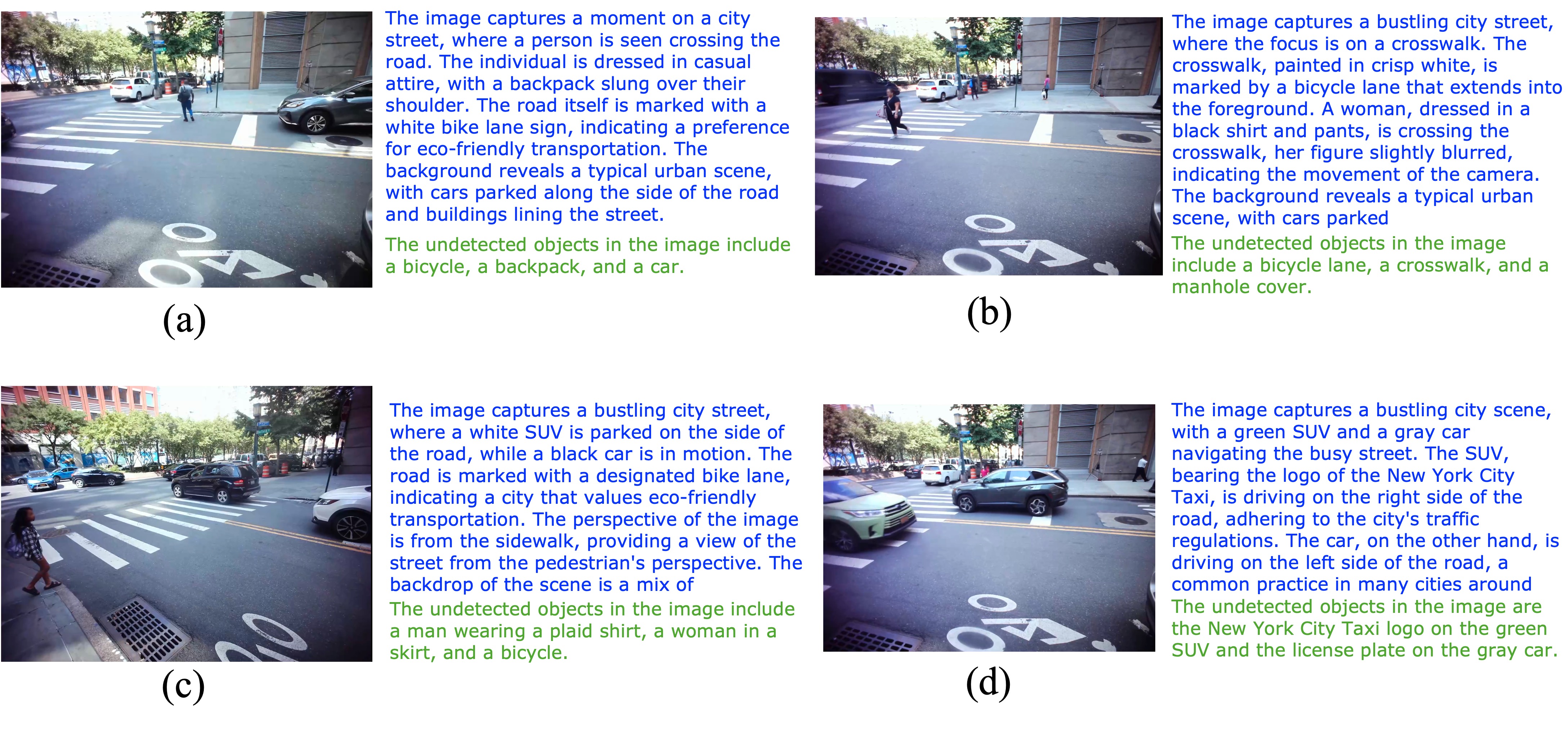}
    \caption{Video clips with descriptions using two prompts with token limit 80 (right camera, referred as base, text shown in blue color) and 32 (left camera text shown in green color). The addition of new information is visible from the text descriptions: (a) ``backpack" is absent in right camera output; (b) ``manhole-cover" can not be detected in the right camera text; (c) ``man wearing a plaid shirt" is not visible from the right camera; and (d) a license plate is detected in the left camera.}
    \label{fig:examples}
\end{figure*}

We consider two Vision-Language Models (VLMs) for experimental analysis: InternVLM-1.8B~\cite{dong2024internlm} and LLAVA-7BV~\cite{liu2023llava}. We run each model independently using the Prompt 1 on each of the video feeds and compute the total ingestion time. We refer this as the \emph{baseline} method. We observe that each model takes approximately 28 to 31 minutes to convert the 46-minute and 44-second long videos to text. Therefore, for a multi-camera setup using 2 cameras, the total ingestion time rises to between 56 minutes and 1 hour, which is time-consuming.

We  also observe the output token statistics for each model in Table~\ref{tab:token-detials}. On average, the InternLM-1.8B model generates between 105 and 113 tokens, while the LLAVA-1.5v2-7B model produces between 142 and 144 tokens. These figures inform the token limits that should be set for text generated by the base camera and other cameras. Following this, in \my, we consider the  right camera as base and convert the video to text using Prompt 1 with token limits ($T_r= $80, and 50 respectively). Subsequently, for the left camera, we utilize Prompt 2 to extract additional details with token limits $T_\ell = 32$. Figure~\ref{fig:examples} displays clips with text from the right camera, limited to 80 tokens (shown in blue), along with additional text from the left camera, limited to 32 tokens (shown in green). It is evident that utilizing textual information from both cameras enhances the accuracy of information compared to relying solely on a single camera, underscoring the value gained from a multi-camera setup.


Table~\ref{tab:ingestion-time-comparison} demonstrates the speed-up in video-to-text conversion time achieved by \my compared to the baseline. For InternLM-1.8B, \my achieves approximately $2\times$ to $3\times$ time reduction, and for LLAVA-7B, \my achieves approximately $3\times$ to $4\times$ time reduction compared with the baseline approach. For InternLM-1.8B (with $T_r=80$ and $T_\ell=32$), adjusting the token limit reduced the conversation time from approximately 56 minutes to 21 minutes. Additionally, the similarity detector further reduced it to 18 minutes. For LLAVA-7B under the same settings, adjusting the token limit reduced the conversation time from approximately 61 minutes to 29 minutes, with the similarity detector further reducing it to 16 minutes. In all these experiments, the similarity threshold has been set to 0.21, as determined by the ablation results described in section~\ref{ablation}.

The addition of new information in subsequent VLM calls as shown in  Figure~\ref{fig:examples} suggests a dissimilarity between the left camera (i.e., camera-n) text and the  right camera (i.e., base) text. A greater degree of text dissimilarity increases the likelihood of uncovering new details about the traffic intersection. To quantify this new information, we employ metrics such as the BERT score~\cite{zhang2019bertscore} and ROUGE score~\cite{lin2004rouge}. Lower BERT and ROUGE scores indicate a higher diversity of information. Our observations, as shown in Table~\ref{tab:variation}, reveal that the scores decrease with the use of targeted prompts and reduced token limits, confirming that text dissimilarity corresponds to the addition of new contextual information.

\begin{table}[!tpb]
    \centering
    \caption{Ingestion information variation among cameras}
    \resizebox{\linewidth}{!}{%
    \begin{tabular}{cccc}
    \toprule
        Model &  Method & BERT score & ROUGE-L \\
        \midrule
         LLAVA-7B & Baseline  & 0.96 & 0.43\\
         \cmidrule{2-4}
         & \my ($T_{r}=80, T_\ell =32$) &  \textbf{0.85} & \textbf{0.26}\\
        \midrule
         InternLM-1.8B & Baseline & 0.95 & 0.42\\
         \cmidrule{2-4}
         & \my ($T_{r}=80, T_\ell =32$) &\textbf{0.85} & \textbf{0.32}\\
         \bottomrule
    \end{tabular}}
    
    \label{tab:variation}
\end{table}

To evaluate the query response capabilities of the Large Language Model (LLM) through textual information generated by \my, we present three complex questions in Table~\ref{tab:street_query}. These questions go beyond simple object identification, such as finding a ``car", and include queries for specific details like ``a white SUV", ``person with backpack", and ``woman with a black bag". These queries require the sophisticated capabilities of the VLM to accurately extract the necessary information. The context being retrieved by the RAG from the text produced by \my based on each query. For all these queries, \my despite reduced text information, successfully generated correct  response.


\begin{table}[!htpb]
    \scriptsize
    \centering
    \caption{Query-response from StreetAware~\cite{streetaware} video dataset.}
    \label{tab:street_query}
    \begin{tabular}{cc|p{5.95cm}}
    \toprule\midrule
    
&\texttt{Q1} & Is there any white SUV? \\
    
    \midrule
     &  RAG Context & 00:07:23 :
The image captures a bustling city street, where a black SUV and a white SUV are parked on the side of the road. The black SUV is positioned on the left side of the image, while the white SUV is on the right. The street is lined with trees and buildings, adding to the urban atmosphere. A crosswalk is also visible in the foreground, indicating a pedestrian-friendly area. The undetected objects in the image are the black SUV and the white SUV.\\
    \midrule
   
& Response &  Yes, there is a white SUV. It is mentioned as being parked on the side of the road.\\
\midrule
\midrule
    
   

&\texttt{Q2} &Is there any person with a backpack?\\
\midrule
     & RAG Context &  00:00:48 :
The image captures a moment on a city street, where a person is seen crossing the road. The individual is dressed in casual attire, with a backpack slung over their shoulder...
. 00:02:05 :
The image captures a moment on a city street, where a pedestrian is seen crossing the road. The pedestrian, dressed in a black shirt and shorts, carries a backpack, suggesting they might be a student or a commuter...
. 00:00:35 :
The image captures a moment on a bustling city street. A man, clad in a white shirt and blue jeans, strides purposefully towards the camera. He carries a backpack, ...
. 00:05:27 :
The image captures a moment on a city street, where a man is crossing the road. He is dressed in a blue shirt and black pants, and he carries a backpack ...\\
    \midrule
   
& Response & Yes, there is a person with a backpack. Multiple instances in the provided context describe individuals carrying backpacks while crossing the street.\\
\midrule
\midrule
&\texttt{Q3} &Is there any woman with a black bag?\\
\midrule
     &  RAG Context & 
00:00:16 :
The image captures a moment on a bustling city street. A woman, dressed in a vibrant red shirt and blue jeans, is seen walking with a black bag. She is crossing the street, adhering to the pedestrian crossing markings. The street is lined with trees, providing a touch of nature amidst the urban setting. Cars are parked along the side of the road, hinting at the. The undetected objects in the image are the woman's black bag and the cars parked along the side of the road.\\

\midrule

& Response &  Yes, there is a woman with a black bag. She is described in the context at 00:00:16:000 as wearing a vibrant red shirt and blue jeans, walking with a black bag while crossing the street.\\
    \bottomrule
    \end{tabular}
\end{table}

%% file: sections/ablation.tex
\section{Ablation Study}\label{ablation}


We conduct an in-depth analysis of the relationship between the similarity threshold, denoted as $\delta$, and the ingestion time required by our system. The similarity threshold $\delta$ plays a crucial role in determining which video clips are processed by the Vision-Language Model (VLM).
Specifically, $\delta$ sets a cutoff point, where only clips with a similarity score lower than $\delta$ are passed to the VLM for further processing.

\begin{figure}[!htbp]
\centering
\begin{subfigure}[b]{0.7\linewidth}
    \includegraphics[width=\linewidth]{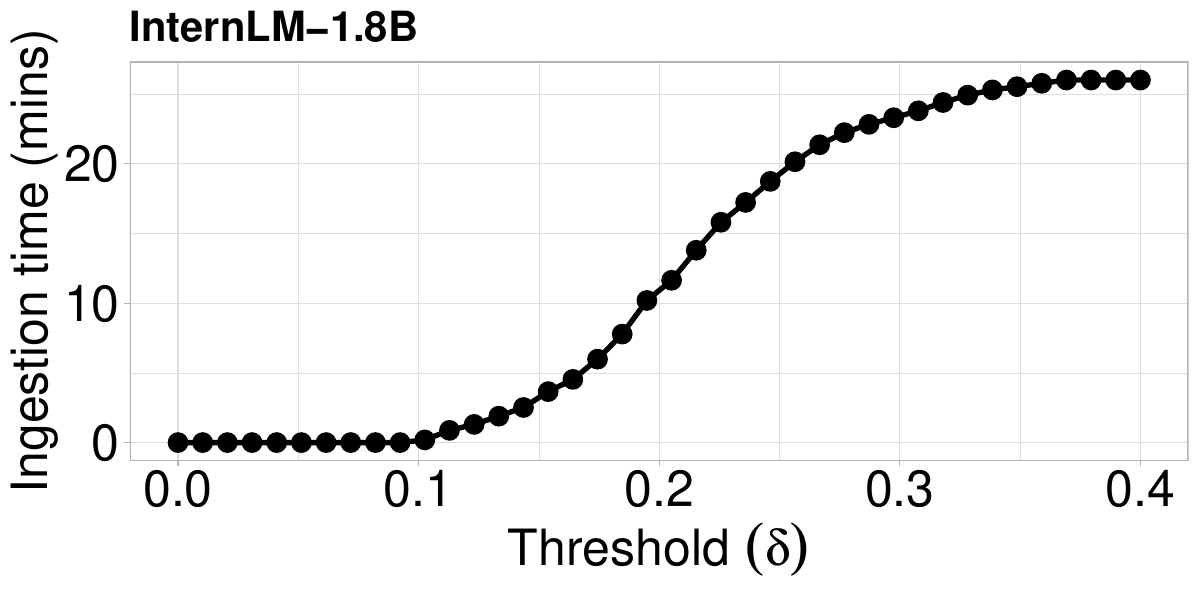}
     \caption{InternVLM-1.8B~\cite{dong2024internlm}}
    \label{fig:sub1}
\end{subfigure}
\begin{subfigure}[b]{0.7\linewidth}
    \includegraphics[width=\linewidth]{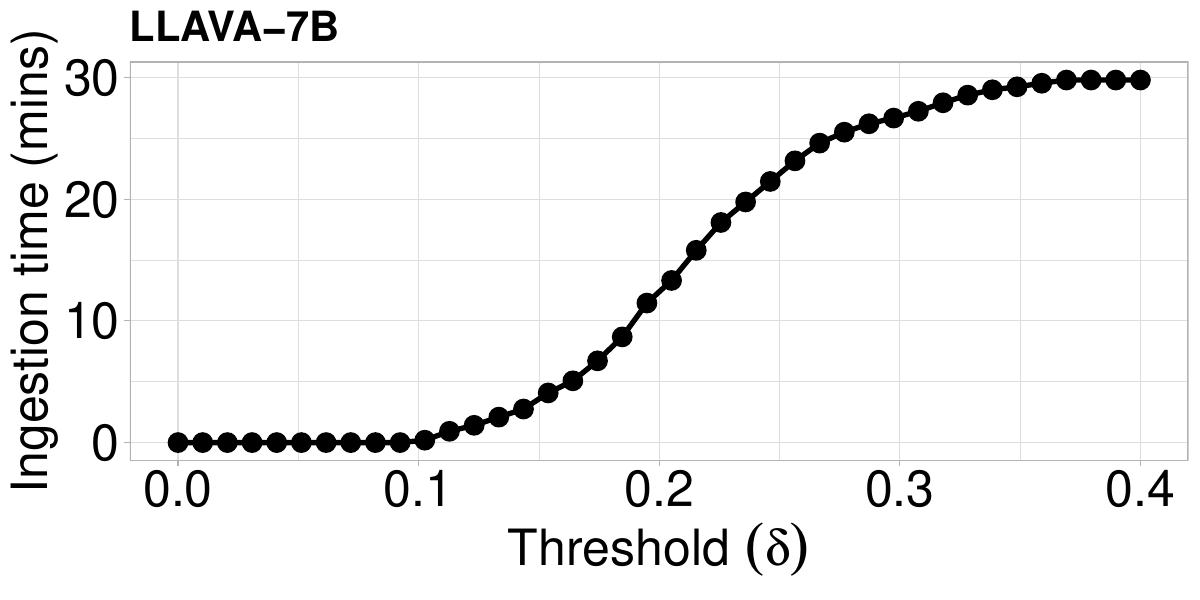}
    \caption{LLAVA-7BV~\cite{liu2023llava}}
    \label{fig:sub2}
\end{subfigure}
\caption{The impact on ingestion time when skipping clips from subsequent cameras during processing with different VLMs using the similarity detector of TrafficLens.}
\label{fig:ablation}
\end{figure}

As illustrated in Figure~\ref{fig:ablation}, our findings reveal a clear trend: as the value of $\delta$ increases, the ingestion time also increases. This is because a higher $\delta$ broadens the range of clips deemed dissimilar enough to require processing by the VLM. Consequently, more clips are subjected to detailed analysis, leading to longer overall ingestion times. On the other hand, when $\delta$ is set to a lower value, the threshold for processing is more stringent, and fewer clips fall below this threshold. As a result, the number of clips requiring VLM processing decreases, thereby reducing the ingestion time. This inverse relationship between $\delta$ and ingestion time is critical for optimizing system performance.


Through extensive analysis of average ingestion times across various threshold values, we have determined that a $\delta$ of 0.21 strikes a balance between processing thoroughness and efficiency. Therefore, this value was selected for all subsequent experiments described in Section~\ref{sec:eval}, as it provides a practical trade-off, ensuring that the system effectively processes necessary clips without incurring unnecessary delays.


%% file: sections/conclusion.tex
\section{Conclusion}

In this paper, we introduce \my, a novel algorithm designed to enhance rapid video-to-text conversion through Vision-Language Models (VLMs) in multi-camera setups at traffic intersections. \my efficiently processes video information by first extracting detailed textual data using a VLM from a base camera with a high maximum token limit. It then applies a lower token limit to gather additional information from other cameras, optimizing the overall video-to-text conversion time. \my also skips text extraction from other cameras when there is a high degree of similarity between the video feeds. This approach ensures quicker data processing while eliminating redundant information extraction. Finally, texts from all cameras covering the same time frames are concatenated and used as a knowledge base within a Retrieval-Augmented Generation (RAG) framework for a Large Language Model (LLM) to generate responses based on user queries. Experimental results demonstrate that \my significantly accelerates video-to-text conversion time without sacrificing information accuracy.